# Investigation of Optimization Techniques on The Elevator Dispatching Problem

Shaher Ahmed, Mohamed Shekha,
Suhaila Skran and Abdelrahman Bassyouny

Department of Mechatronics Engineering, Faculty of Engineering and
Materials Science, The German University in Cairo, Egypt

## Abstract

*In the elevator industry, reducing passenger journey time in an elevator system is a major aim. The key obstacle to optimising elevator dispatching is the unpredictable traffic flow of passengers. To address this difficulty, two main features must be optimised: waiting time and journey time. To address the problem in real time, several strategies are employed, including Simulated Annealing (SA), Genetic Algorithm (GA), Particle Swarm Optimization Algorithm (PSO), and Whale Optimization Algorithm (WOA). This research article compares the algorithms discussed above. To investigate the functioning of the algorithms for visualisation and insight, a case study was created. In order to discover the optimum algorithm for the elevator dispatching problem, performance indices such as average and ideal fitness value are generated in 5 runs to compare the outcomes of the methods. The goal of this study is to compute a dispatching scheme, which is the result of the algorithms, in order to lower the average trip time for all passengers. This study builds on previous studies that recommended ways to reduce waiting time. The proposed technique reduces average wait time, improves lift efficiency, and improves customer experience.*

## Keywords

*Stochastic Optimization, Elevator Dispatching Systems, Meta-Heuristics Optimization Techniques.*

## 1. Introduction

Elevators are designed to transport passengers safely, comfortably, and efficiently from their initial floors to their destinations. The use of elevators has been increasing throughout the years, as more construction is happening. An important concern for elevator producers is the importance of elevator efficiency, the reduction of passenger waiting time, and reducing the time to the destination. People prefer to use elevators rather than stairs to reach the upper levels of a building. People experience various challenges with the traditional elevator management system with just one operating lift, such as long wait times for the lift and long trip times since several passengers share a single lift and installing another lift might be highly expensive.

The Elevator Dispatching Problem (EDP) has become a popular research topic in both academia and industry as it is an excellent example of a stochastic optimal control problem of economic importance that is too large to solve using traditional techniques such as dynamic programming. Several studies use optimization techniques to improve the performance of the controller or elevator. The research gap of this research is the need for a comparative study to assess the optimality and efficiency of the elevator dispatching system. Several studies were conducted on the EDP offering multiple solutions to this problem, such as the use of algorithms and dispatching scheduling, the use of multiple elevators, and finally, the assistance of optimization techniques. These solutions were introduced for the purpose of optimizing the journey time, to achieve higher efficiency of lift transportation.

This paper gives a comprehensive literature review of several strategies used to improve the efficiency of the Elevator Dispatching Problem. Optimization Techniques algorithms are some of the strategies used to optimise elevator operation.

The rest of this paper is organized as follows: in Section 2, we review appropriate literature on stochastic and dynamic vehicle routing as well as on elevator dispatching. We formulate the EDP in Section 3. Section 4 describes and presents the algorithms that were utilised to solve the EDP. Section 5 shows the output of the algorithms which is the elevator routes to be taken. Finally, in Section 6, we finalise this study and outline future research options.

## 2. LITERATURE REVIEW

M. Latif, M. Kheshaim, and S. Kundu proposed EDP algorithms such as Estimated Time to Dispatch (ETD), Destination Control System (DCS), and Compass Plus Destination System (CDS) in 2016. The ETD algorithm uses destination-based dispatching system, which asks users to register their destination floor, thus assigning them to an elevator car. The DCS algorithm groups users according to their registered destinations and integrates artificial intelligence, travel forecasting, fuzzy logic and genetic optimization and continuously monitors the traffic behaviour at all times to avoid malfunctions. The CDS algorithm introduces the conventional smart grouping concept. Passengers with similar destination is assigned the same elevator car, and elevators are assigned to specific floors, leading to fewer stops being made thus reducing time to reach the destination. [1]

M. Sale and V. Prakash mentioned in 2019 that the optimization strategies of Elevator Dispatching Systems may be classified as fuzzy based algorithms, genetic algorithms, neural network algorithms, artificial intelligence algorithms, and zoning algorithms. The Fuzzy Algorithm reduces waiting time and improves the service quality and performance however it is a robust method. The memory may face faults which affects the performance of fuzzy controllers. The Artificial intelligence, Neural Network have a very optimal dispatching policy but needs expert experiences and training efforts to achieve good results and are difficult to understand and implement. Zoning algorithms have reduced energy consumption and have an optimal dispatching policy but it becomes robust and less flexible in heavy traffic. [2]

In a paper conducted by Sorsa, J., et al. that studied the EDP by solving a snapshot optimization problem. Passengers register their destination floors in elevator lobbies, after which the elevator group control system (EGCS) completes the assignment immediately and cannot be changed afterwards. A Poisson and a geometric Poisson process are used to describe stochastic requests and consumers. This leads to improved forecasting accuracy than the Poisson process and several scenarios that closely match the variable realisations. As a result, the scenarios may be utilised as the foundation for a robust EDP that minimises both a passenger service quality criterion and its volatility owing to variable demand. [3]

A. So and L. Al-Sharif, considered the Idealized Optimal Benchmark concept used in various engineering projects like in thermodynamics to produce the most efficient heat engine possible, so this concept can be used in the traffic elevator design by calculating the value of the round-trip time which is the time taken by the elevator to complete the task of transporting and picking up the passenger to the desired destination to produce the maximum performance of an elevator system. Number of stops and highest reversal floor are important parameters that should be at the minimum possible values and these values can be done by specific formulae. The number of the elevators are sequenced to the virtual sectors in the building to equalize the handling capacities of the sectors. [4]

An EGCS is one of the important matters in vertical transportation systems in buildings, it is required to tackle some optimization problems. It dispatches the elevators under its control to perform each passenger's call. While the elevators are fulfilling their present calls, more passengers may emerge. This makes the control problem dynamic. Its solution defines complete elevator route by minimizing passengers' waiting or journey times. Janne Sorsa [5] considered the EDP as a bi-level, dynamical, stochastic, multi-objective optimization problem. Since the EDP is a difficult optimization problem, its solving time using an exact algorithm increases exponentially with respect to the problem's size. The main effective solution approach shall be based on a genetic algorithm suitable for real-time optimization. Tartan et al. [6] used a Genetic Algorithm that has attracted researchers to study the suggested optimization of waiting time. The suggested solution minimises average passenger waiting time and employs a simpler encoding methodology, resulting in computational cost efficiency.

In 2018, A. Vodopija, J. Stork, T. Bartz-Beielstein, and B. Filipic discussed the S-Ring based elevator and its operation on the principle of a normal ring road. It is usually used in the elevator group control. The system works on the principle of neural network, at each state it is checked whether a new customer arrived. The controller then decides whether the elevator car should stop or continue to the next state if the current state is an active elevator state. Finally, the active state indicates whether the customer has been served or not. [7]

In addition, in 2019, Bapin, Y., and Zarikas, V. created an algorithm that employs an image collecting and processing system to make use of information regarding passenger group sizes and waiting times. The study aimed for the construction of a decision engine capable to control the elevators actions in order that improves user's satisfaction. The data is utilised by the probabilistic decision-making model to perform Bayesian inference and update variable parameters. The results demonstrated that the suggested algorithm demonstrated the intended behaviour in 94% of the circumstances examined. [8]

Z. Yang and W. Yue investigated Elevator Traffic Pattern Recognition Using Fuzzy BP Neural Network with Self-Organizing Map (SOM) Algorithm in 2017. SOM is a form of unsupervised learning network method that operates on the premise of grouping comparable inputs on the same output in order to find a better clustering centre after a sufficient number of repetitions. Alongside with the fuzzy logic theory that provides a working algorithm to solve the control system. With the help of the neural network technique that features a stronger self-learning ability which is helpful when dealing with a nonlinear and uncertain dynamics and changes of the elevator traffic. When placing both theories together, each one covers for the drawbacks of the other to obtain the best results. [9]

M. Sale and V. Prakash discussed the design of an elevator system has many factors that should be took in consideration like managing time and money. EGCS is responsible for making several efficient elevators in carrying passengers. EGCS is a solution by reducing the waiting time during the peak time where all passengers need to use the elevator, this system uses fuzzy algorithm. Fuzzy system is a problem-solving rule based. It mainly consists of four blocks: fuzzy rule for applying on data, fuzzifier to accept input values, DeFuzzifier for generating output and inference engine for operations. There are two modules in the system: Hall call module which is responsible for sending the calling floor id to the controller and car call module which is responsible for sending the destination floor id to the controller. [10]

## 3. PROBLEM STATEMENT AND FORMULATION

The main objective of this problem is to provide passengers with the shortest waiting and destination time. Thus, this can be considered as multi-objective optimization problem. Both parameters can be combined into one parameter; journey time, thus, the objective is to minimise the average journey time for all passengers. The model should have some parameters that should be specified. These parameters are then used to evaluate the two main equations for the waiting and destination time.

Table 1. Table of Fixed parameters of the elevator

| Fixed Parameters | Time |
|---|---|
| Opening Time (OT) | 2 Seconds |
| Closing Time (CT) | 2 Seconds |
| Passengers Load Time (PLT) | 5 Seconds |
| Between Floors (BFT) | 5 Seconds |

$$\text{Load Time } (L_T) = O_T + C_T + PL_T \tag{1}$$
$$N_1 = |F_N - C_N| \tag{2}$$
$$N_2 = |F_N - D_F| \tag{3}$$

$N_1$ is defined to be the number of floors between the passenger and the elevator, $F_N$ is the floor the elevator is at and $C_N$ is the call floor number. $N_2$ is defined to be the number of floors between the passenger and the elevator, and $D_F$ is the destination floor number.

$C_B$ is defined to be the number of calls made before the current passenger, $C_A$ is defined to be the number of calls made after the current passenger and $D_B$ is defined to be the number of drops offs before the current passenger.

Thus, the first objective function equation for calculating the waiting time of a passenger is defined to be:

$$\text{Waiting Time } (W_T) = BF_T * N_1 + L_T * C_B + L_T * D_B + O_T \tag{4}$$

Thus, the second objective function equation for calculating the destination time of a passenger is defined to be:

$$\text{Destination Time } (DT) = C_T + L_T * C_A + BF_T * N_2 + L_T * D_B + O_T \tag{5}$$

Finally, the objective of the EDP is the summation of equations (4) and (5):

$$J_T = D_T + W_T \tag{6}$$

Where $J_T$, $D_T$, and $W_T$ denote journey, destination and waiting times respectively.

The average journey time is the output of the algorithms, also known as the fitness value, which is the parameter desired to be minimized and it presents the average journey time for all passengers. Where n represents the number of passengers.

$$\text{Average Journey Time} = \sum_{1}^{n} JT(n)/n \qquad (7)$$

In order to compare the algorithms, a case study was created to analyse the functioning of the algorithms and to produce a plausible answer. The decision variables of the EDP are the floor numbers, which is represented in an array, presenting the route that the elevator will follow in order to reach the average journey time or for all passengers.

A building was selected for a case study serviced by a single elevator. The floors of the building start from the ground floor, and end in the 20th floor (20). Thus, the building has 21 floors in total. The case is studied on 10 passengers in total. The initial floor of the elevator is the 4th floor.

Table 2. Case study

|  | **Case Study Scenario** |
|---|---|
| Passenger 1 | 5 → 9 |
| Passenger 2 | 6 → 7 |
| Passenger 3 | 3 → 15 |
| Passenger 4 | 11 → 0 |
| Passenger 5 | 20 → 8 |
| Passenger 6 | 10 → 17 |
| Passenger 7 | 13 → 19 |
| Passenger 8 | 1 → 14 |
| Passenger 9 | 16 → 2 |
| Passenger 10 | 18 → 12 |

## 4. METHODOLOGY

In this paper, four optimization techniques are discussed and implemented on the EDP. This section presents the algorithms; Simulated Annealing (SA), Genetic Algorithm (GA), Particle Swarm Optimization Algorithm (PSO), and Whale Optimization Algorithm (WOA).

The EDP is a permutation problem; thus, the output of the algorithms will be the route that the elevator must follow to reach the desired objective function.

### 4.1. Trajectory based Stochastic Technique

#### 4.1.1. Simulated Annealing (SA)

One of the most used heuristic approaches for tackling optimization problems is the SA algorithm. The annealing technique describes the ideal molecular configurations of metal particles in which the potential energy of the mass is reduced, and it refers to progressively cooling the metals after they have been exposed to high heat. In general, the SA algorithm employs an iterative movement based on a changeable temperature parameter, simulating the annealing transaction of metals. A basic optimization technique compares the outputs of the objective functions running with the current and neighbouring points in the domain repeatedly, so that if the neighbouring point produces a better result than the current one, it is saved as the base solution for the next iteration. Otherwise, the algorithm quits the operation without attempting to search the larger area for better results.

Parameters $T_i$ (initial temperature), $T_f$ (final temperature) and $i_m$ (maximum number of iterations) are first initialized. The change in energy $\Delta E$ is then computed for the feasible solution. If this value is less than 0, the solution is then accepted and the best solution is also updated. If the solution is greater than 0, this solution is first evaluated by 2 parameters, $r$ and $P$. $r$ is a random number initialized between 0 and 1.

If the value of $r$ is less than $P$, the solution will be accepted, updating the current value, however, the best solution so far is not updated and the temperature value is then updated. This loop is repeated while the stopping criteria has not been met; the temperature value is greater than the $T_f$ value, and as long as the iterations is less than the $i_m$ value. [11]

```
• Decision Variables: x = x_1, x_2, ..., x_p
• Objective Function: f(x)

Initialize x_0, T_0, T_f, i_max, x*, f*
While T_cur > T_f and i ≤ i_max:
    Repeat for n_T (iterations per temperature):
        Generate random new feasible solutions: x_new = x_cur + rand
        Compute change in energy: ΔE = Δf = f(x_new) − f(x_cur)
        If ΔE < 0:
            Accept better solution: f(x_new)
        else if r < p = e^(-ΔE/T) ( r is a random number 0~1):
            Accept solution: x_cur = x_new
        If f(x_cur) < f*:
            Update best reached solution: x*, f* = x_cur, f(x_cur)
    Update temperature according to cooling schedule.
    Update iteration counter: i = i + 1
Output best reached solution x*, f*
```

Fig. 1. Pseudo-code of the SA algorithm

Table 3. Used Parameters on the SA algorithm

| Parameter | Value |
|---|---|
| Population size | 1 |
| Initial Temperature | 200 |
| Final Temperature | 0.01 |
| No. of iterations | 100 |
| No. of iteration per temperature | 1 |

## 4.2. Population based Stochastic Technique – 1/2

In the population based stochastic techniques, the loop is operating while the stopping criteria has not been reached, which are: 1) current number of iterations is less than the $i_m$ value, 2) when the objective function value has reached a certain pre-defined value 3) when there has been no improvement in the population over a number of iterations.

**4.2.1. Genetic Algorithm (GA)**

Genetic algorithms are randomised search algorithms designed to mimic the mechanics of natural selection and natural genetics. Genetic algorithms work on string structures, which, like biological structures, evolve over time according to the rule of survival of the fittest through a randomised yet organised information exchange. Thus, in every generation, a new set of strings is created, using parts of the fittest members of the old set. At first, the coding to be used must be defined. Then using a random process, an initial population of strings is created. Following that, a series of operators is employed to take this initial population and construct consecutive populations that should improve with time. The genetic algorithms' core operators are reproduction, crossover, and mutation. Reproduction is a process that is based on each string's goal function (fitness function). This objective function identifies how "good" a string is. Thus, strings with higher fitness value have bigger probability of contributing offspring to the next generation. Crossover is a process in which members of the last population are mated at random in the mating pool. So, a pair of offspring is generated, combining elements from two parents (members), which hopefully have improved fitness values. Mutation is the random (but infrequent) modification of the value of a string position. Mutation is, in reality, a random walk across the coded parameter space. Its goal is to prevent essential information contained inside strings from being lost prematurely.

The Genetic Algorithm (GA) is a population based stochastic technique. $X_{prev}$ is the previous values of $X$ from the previous loop.

Parameters $m$ (population size) of size 4, $pe$ (Elite members survived from $X_{prev}$) 1 Elite member, $pc$ (Cross-over best members survived from $X_{prev}$) 2 Cross-over member, and $pm$ (mutant worst members survived from $X_{prev}$) 1 Mutant member are first initialized. A new set of feasible solutions (chromosomes) are generated, and depending on the fitness value of each chromosome, it is categorized to either elite, cross-over or mutant. For the cross-over and mutant members, operations must occur on these members in order to generate a new child from its corresponding parents. Davis-order crossover is used for the cross over members and swap mutation is used for the mutant members. Post the operation, the fitness values are then computed once again for all the chromosomes and the iteration is repeated until reaching the stopping criteria. [12]

- Decision Variables: $\underline{x} = x_1, x_2, \ldots, x_p$
- Objective Function: $f(\underline{x})$
- Initialize $\underline{x}, i_{max}, m, p_e, p_c, p_m, x^*, f^*$

While **_Stopping Criteria_** Ex. ($i \leq i_{max}$):
    Generate random new population with (m) chromosomes $x_{new}$:
        - $p_e$ Elite members are survived from $x_{prev}$
        - $p_c$ Cross-over best members from $x_{prev}$
        - $p_m$ mutate worst members from $x_{prev}$
    Compute fitness value: $f(x_{new})$
    Update iteration counter: $i = i + 1$
Output best reached solution $x^*, f^*$

Fig. 2. Pseudo-code of the GA algorithm

Table 4. Used Parameters on the GA algorithm

| Parameter | Value |
|---|---|
| Population size | 4 |
| No. of iterations | 100 |
| Elite member | 1 |
| Crossover member | 2 |
| Mutant member | 1 |

**4.3. Population based Stochastic Technique – 2/2**

The PSO and WOA deals with arithmetic problems and the EDP is a permutation problem, thus, a space transformation must be done to use this algorithm in order to use the algorithms on the EDP.

**4.3.1. Particle Swarm Optimization Algorithm (PSO)**

The particle swarm optimization (PSO) is a randomized, population-based optimization method that was inspired by the flocking behaviour of birds or fish schooling. In PSO, each single solution is a "bird" in the search space. We call it a "particle". A swarm of these particles moves through the search space to find an optimal position. PSO is initialized with a group of random particles (solutions) and then searches for optima by updating generations. During every iteration, each particle is updated by following two "best" values. The first one is the position vector of the best solution (fitness) this particle has achieved so far. The fitness value is also stored. This position is called *pbest*. Another "best" position that is tracked by the particle swarm optimizer is the best position, obtained so far, by any particle in the population. This best position is the current global best and is called *gbest*. After finding the two best values, the position and velocity of the particles are updated.

Parameters $m$ (Population size), $w$ (Inertia weight), $c1$ and $c2$ (cognitive factor and social factor) and $i_m$ (maximum number of iterations) are first initialized. Initial Velocity is an array, initialized at 0 for all elements. New population of random feasible solutions are generated, and for every particle, the velocity is first updated using its equation, then the position is then updated. The fitness values of all populations are then computed followed by updating the personal and the neighbourhood best. A previous study [13] is followed to convert the solution representation of the PSO to a permutation one. This modified algorithm is implemented using the smallest place value rule, which finds the permutations between continuous positions $X_{kj}$. A new sequence vector is generated and represented by $S_{kj}$ The values inside the continuous position's ($X$) vector are arranged in their ascending order, then the reflecting indices of this order are added in the $S$ vector by the order of presence the corresponding arithmetic value in the original (unchanged) $X$ vector. [14]

- Decision Variables: $\underline{x} = x_1, x_2, \ldots, x_p$
- Objective Function: $f(x)$
- Initialize $X_0, V_0, i_{max}, m, w, c_1, c_2, x^*, f^*$

While **Stopping Criteria** Ex. ($i \leq i_{max}$):
    Generate random new population with (m) particles $X_{i+1}$.
    For each particle $x_i$ in current population $X_i$:
        - Velocity Update:
            $r_1, r_2 = random\ number\ vectors\ (same\ size\ as\ x_i)$
            $v_{i+1} = w\ v_i + c_1 r_1 (Pbest_i - x_i) + c_2 r_2 (NBest_i - x_i)$
        - Position Update:
            $x_{i+1} = x_i + v_{i+1}$
    Compute fitness value: $f(X_{new})$
    Update personal best: $Pbest_{i+1} = x_{i+1}\ if\ f(x_{i+1}) < f(Pbest_i)$
    Update neighborhood best: $Nbest_{i+1} = \underset{N}{\mathrm{argmin}}(f(x_{i+1}))$
    Update iteration counter: $i = i + 1$
Output best reached solution $x^*, f^*$

Fig. 3. Pseudo-code of the PSO algorithm

Table 5. Used Parameters on the PSO algorithm

| Parameter | Value |
|---|---|
| Population size (*m*) | 4 |
| No. of iterations | 100 |
| inertia weight (*w*) | 0.792 |
| Cognitive and Social factor (*c1, c2*) | 1.4994 |

### 4.3.2. Whale Optimization Algorithm (WOA)

Whale Optimization Algorithm (WOA) is a recently proposed optimization algorithm mimicking the hunting mechanism of humpback whales in nature. The WOA algorithm starts with a set of random solutions. At each iteration, search agents update their positions with respect to either a randomly chosen search agent or the best solution obtained so far. The '*a*' parameter is decreased from 2 to 0 in order to provide exploration and exploitation, respectively. A random search agent is chosen when *A*>1, while the best solution is selected when *A*<1 for updating the position of the search agents. Finally, the WOA algorithm is terminated by the satisfaction of a termination criterion.

5 Parameters are being updated at the beginning of each iteration, which are *a, A, C, l* and *p*. *l* and *p* are random numbers generated from 0 to 1, while the remaining parameters are updated using their corresponding equations.

Depending of the values of the above parameters, the corresponding operation is executed. If P < 0.5 and *A* < 1; the Encircling Prey method is used. If P < 0.5 and *A* >= 1; the Search for Prey method is used. If *P* >= 0.5; the Spiral Update method is used. The fitness value of all agents is calculated and the agent that has the best fitness value is compared to the overall reached best agent, and is updated if it is better than it. [15]

A space transformation used to implement a permutation problem using the WOA. The great value priority rule was used to transform the arithmetic solution to a permutation solution. This is done by sorting the elements of the solution in descending order, then taking the indices of these elements and use them as the solution of the WOA. Also, in order to improve the performance and quality of solutions, local search was applied by selecting 2 elements at random to swap if a better tour is found.

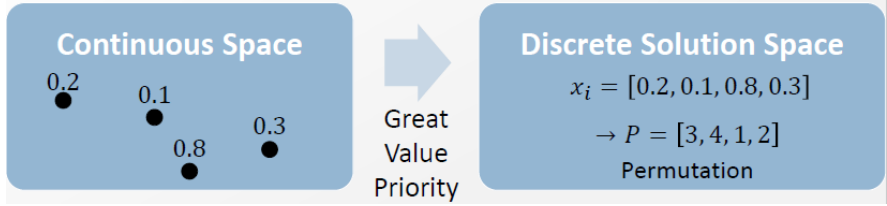

Fig. 4. Example of the great value priority transformation

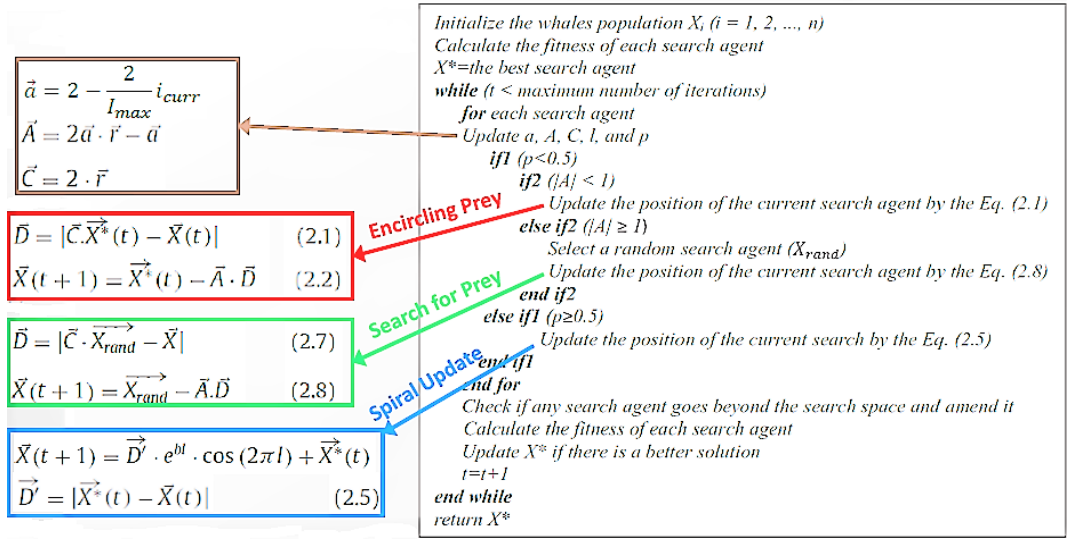

Fig. 5. Pseudo-code of the WOA algorithm

Table 6. Used Parameters on the WOA algorithm

| Parameter | Value |
|---|---|
| Population size ($m$) | 4 |
| No. of iterations | 100 |

## 5. RESULTS AND DISCUSSION

The case study is tested on all four implemented algorithms. Each case study's result is plotted on a convergence plot graph, where the X axis is the number of iterations and the Y axis presents the fitness values.

For the studied case implemented on the SA, the simulations best solution (route to be taken) is [4,1,3,5,16,10,9,14,20,13,6,18,15,11,12,19,7,8,17,2,0]. The average journey time of the optimized solution is approximately 472.3 seconds.

For the studied case implemented on the GA, the simulations best solution (route to be taken) is [4,3,1,5,6,7,9,10,11,13,14,15,16,17,18,19,20,12,8,2,0]. The average journey time of the optimized solution is approximately 222.3 seconds.

For the studied case implemented on the PSO, the simulations best solution (route to be taken) is [4,20,18,13,11,10,12,8,1,16,0,2,3,19,5,17,15,14,6,9,7]. The average journey time of the optimized solution is approximately 553.3 seconds.

For the studied case implemented on the WOA, the simulations best solution (route to be taken) is [4,5,9,10,16,18,12,1,13,2,14,19,6,20,17,3,8,15,11,7,0]. The average journey time of the optimized solution is approximately 541.3 seconds.

It is observed, that with the increase of the number of iterations, the solution converges to the best possible fitness value of the algorithm, with respect to the case study. It is also observed during trial runs, that the more the passengers in a case study, the more iterations needed to reach an optimal solution, as there are several combinations of solutions possible.

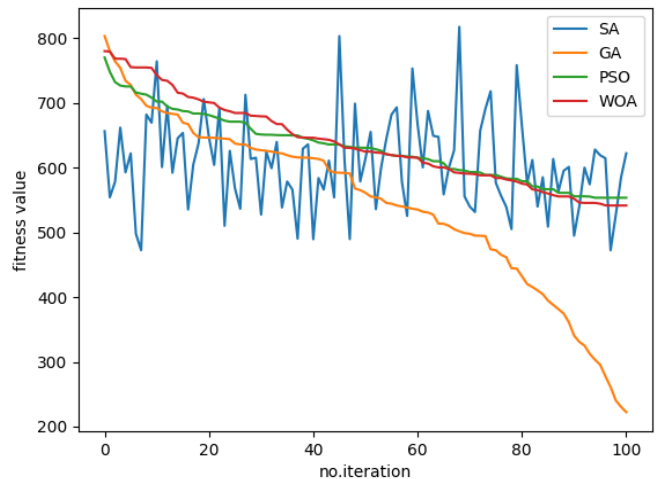

Fig. 6. Convergence Plot of SA (Blue), GA (Orange), PSO(Green) and WOA (Red) respectively

## 5.1. Performance Comparison

Table 7. Performance Comparison of the algorithms

| Performance Index | SA | GA | PSO | WOA |
|---|---|---|---|---|
| Avg. Fitness value in 5 runs | 531.4 s | 279.1 s | 600 s | 620 s |
| Optimal Fitness Value | 472.3 s | 222.3 s | 553.3 s | 541.3 s |

As per the data shown in Table 7 and Figure 6, we can see that the genetic algorithm has the best fitness value over the period of 100 iterations, while the simulated annealing algorithm is fluctuating between the fitness values. As for the PSO and WOA, the algorithms are reaching the optimal solution over the number of iterations and it is expected that if the number of iterations is increased, based on the graphs of both algorithms, it will reach an optimal solution. Both algorithms are affected by the space transformation, thus this is why both algorithms are preforming similarly and their fitness values are not so far from the GA and SA. The GA performs best over the remaining algorithms and over 5 runs as well, while the WOA preforms worst. This could be caused by the random variables that are in algorithm.

## 6. CONCLUSIONS AND FUTURE RECOMMENDATIONS

In conclusion, because of the growth of large buildings and the increased demand for more sophisticated elevators, elevator dispatch systems are always being enhanced and evolving. These upgraded methods are designed to reduce wait times.

This current research work evaluates the usage of four different optimization algorithms to provide a solution for the Elevator Dispatching Problem. The algorithms, Simulated Annealing (SA), Genetic Algorithm (GA), Particle Swarm Optimization Algorithm (PSO), and Whale Optimization Algorithm (WOA) are implemented using Python programming language and tested on a case to solve the Elevator Dispatching Problem in order to improve lift efficiency and provide a better user experience. To evaluate and test the performance of the four algorithms, performance indices are computed such as average and optimal fitness value in 5 runs to consider which of the algorithms is best to use in the industry. Based on the testing and findings, the Genetic Algorithm performed better than other algorithms. Our future recommendation is to use the Genetic Algorithm (GA) in the practical and dynamic elevator dispatching systems. We also highly recommend using adaptive optimization techniques. Further research is required to develop, implement and evaluate the use of multiple elevators. We advise that this study be improved further by collecting elevator traffic data throughout the course of a complete working day, running the algorithms at different periods of the day using different case studies, and taking the average of the results. A limitation that could improve the study is the inclusion of same floor calls from different passengers.

### ACKNOWLEDGEMENTS

The authors thank and acknowledge the assistance and the usage of lecture notes provided by *Prof. Dr. Omar M. Shehata*, *M.Sc Eng. Shaimaa El Baklish*, and *M.Sc Eng. Catherine M. Elias* at the Department of Mechatronics Engineering of The German University in Cairo. The authors would like to thank the anonymous reviewers for their insightful remarks that helped enhance this research work.

## AUTHORS

**Shaher Ahmed** is the first author and the lead researcher of this paper. He is currently pursuing a M.Sc degree focused in Industrial Automation, Machine Learning and Wireless Sensor Networks at The German University in Cairo's Department of Mechatronics Engineering. He graduated from the same university with a B.Sc in Mechatronics Engineering. His primary areas of interest include Optimization, Machine Learning, and Industrial Automation.

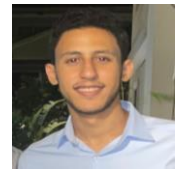

**Mohamed Shekha** graduated from The German University in Cairo with a B.Sc in Mechatronics Engineering. His major areas of interest include Automation and Intelligent Systems, particularly in the Automotive industry.

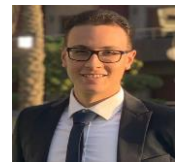

**Suhaila Skran** holds a bachelor's degree from the German University in Cairo, Egypt. She has a Bachelor of Science degree in Mechatronics Engineering. "System Identification of Active Quarter Car Model Using Deep Learning" was the title of her thesis. She works full-time at Valeo Egypt as an Automotive Embedded Software Engineer.

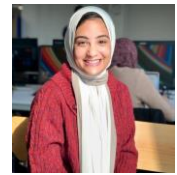

**Abdelrahman Bassyouny** holds a B.Sc in Mechatronics Engineering from the German University in Cairo.

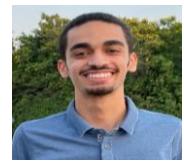